\documentclass[11pt]{article}

\usepackage[preprint]{acl}
\usepackage{times}
\usepackage{latexsym}
\usepackage[T1]{fontenc}
\usepackage[utf8]{inputenc}
\usepackage{microtype}
\usepackage{inconsolata}
\usepackage{graphicx}
\usepackage{booktabs}
\usepackage{amsmath}
\usepackage{amssymb}
\usepackage[table]{xcolor}
\usepackage[most]{tcolorbox}
\usepackage{xspace}

\DeclareMathSizes{7.4}{7.4}{5.18}{5}

\newcommand{\method}{\textsc{TIR-EC}\xspace}
\definecolor{accent}{HTML}{2C6593}
\definecolor{teal}{HTML}{367F76}
\newtcolorbox{takeaway}{enhanced,colback=teal!5,colframe=teal!55,boxrule=.45pt,arc=1mm,left=1.5mm,right=1.5mm,top=1mm,bottom=1mm,before skip=4pt,after skip=3pt,fontupper=\small}
\definecolor{tirreason}{HTML}{174A7E}
\definecolor{tirtool}{HTML}{087F5B}
\definecolor{tirobservation}{HTML}{8A4B08}
\definecolor{tiranswer}{HTML}{6B3FA0}
\newcommand{\reasonfmt}[1]{\textcolor{tirreason}{#1}}
\newcommand{\toolcallfmt}[1]{\textcolor{tirtool}{#1}}
\newcommand{\observationfmt}[1]{\textcolor{tirobservation}{#1}}
\newcommand{\answerfmt}[1]{\textcolor{tiranswer}{#1}}
\newcommand{\blackcircled}[1]{%
  \tikz[baseline=(num.base)]{
    \node[
      circle,
      fill=black,
      text=white,
      inner sep=0pt,
      minimum size=1em,
      font=\sffamily\bfseries
    ] (num) {#1};
  }%
}
\usepackage{booktabs,graphicx,xcolor,colortbl}
\definecolor{bestblue}{HTML}{B4CDED}
\definecolor{secondblue}{HTML}{DCE8F7}

\title{\texorpdfstring{\raisebox{-0.2em}{\includegraphics[height=1.3em]{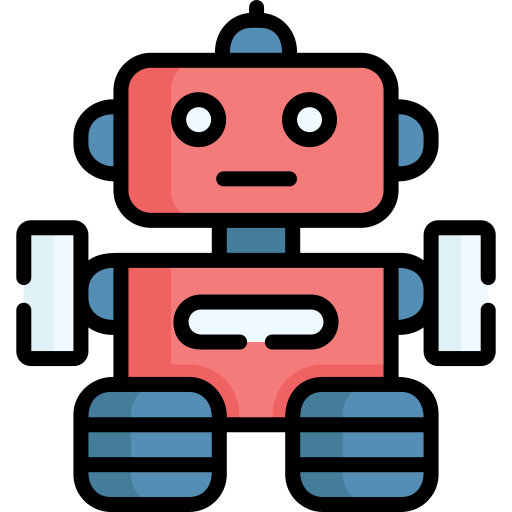}}\hspace{0.3em}}{}\method: Incentivizing Tool-Use Reinforcement Learning for Post-Arrival Emergency Care Prediction}

\author{
  {\normalsize\bfseries Pengfei Hu\textsuperscript{1}\thanks{Equal contribution.}\enspace
  Chenwei Wu\textsuperscript{2}\footnotemark[1]\enspace
  Qiao Jin\textsuperscript{3}\enspace
  Jun Deng\textsuperscript{4}\enspace
  Zhiyong Lu\textsuperscript{3}\enspace
  Liyue Shen\textsuperscript{2}\thanks{Corresponding authors.}\enspace
  Yue Ning\textsuperscript{1}\footnotemark[2]}\\
  {\small\textsuperscript{1}Stevens Institute of Technology\quad
  \textsuperscript{2}University of Michigan, Ann Arbor}\\
  {\small\textsuperscript{3}National Institutes of Health\quad
  \textsuperscript{4}Yale University}\\
  {\small\texttt{\{phu9,yue.ning\}@stevens.edu}\quad
  \texttt{\{chenweiw,liyues\}@umich.edu}}\\
  {\small\texttt{\{qiao.jin,zhiyong.lu\}@nih.gov}\quad\texttt{jun.deng@yale.edu}}
}

\begin{document}
\maketitle

\begin{abstract}
Large language models (LLMs) can reason over both structured measurements and free-text records, which makes them attractive for emergency department (ED) prediction. However, most existing LLMs still fall short of conventional ML methods like trees on EHR-based prediction tasks. Our controlled analysis traces this gap to two weaknesses: LLMs do not select informative features, and they misjudge what numerical values imply for risk even when given the right features. We propose TIR-EC, a tool-integrated reinforcement learning framework that addresses both. TIR-EC exposes EHR feature groups as tools the LLM can call. It trains the LLM with rewards from a gradient-boosted tree teacher, which provides a reference set of informative tools and a graded risk target. TIR-EC also distills task-specific clinical skills into the model weights through a reverse curriculum that gradually removes the skills from the prompt during RL process. Evaluation on three clinical prediction tasks shows that our 4B agent achieves the highest macro F1 of all compared models, outperforms conventional ML baselines, and is competitive with much larger frontier LLMs. Further analysis shows that TIR-EC selects more useful evidence with fewer tool calls and combines measurements more effectively.
\end{abstract}

\section{Introduction}
\label{sec:introduction}
The emergency department (ED) is a high-stakes, time-critical scenario where predicting outcomes such as hospital admission, clinical deterioration, and return visits from electronic health records (EHRs) can support timely decision-making~\citep{xie2022edbenchmark,chen2023mcbec,wu2025towards}. 
Unlike conventional machine learning models, large language models (LLMs) can jointly reason over structured EHR data and free-text clinical notes, like how a clinician proceeds, and therefore its application in ED have attracted growing research attention~\citep{shi2024ehragent,alcaraz2025mdsed,kruse2025temporal}.


However, recent studies show that most LLMs, both general-purpose and medical, still underperform simple and cheap conventional models such as gradient-boosted trees on EHR-based clinical prediction in spite of model scaling, prompting, or fine-tuning~\cite{chen2026clinicalbench,sanatizadeh2026generalization}. To understand \textit{why LLMs underperform on EHR-based clinical prediction, and more importantly, how they could be improved}, we run a controlled preliminary analysis on ED encounters comparing general and medical LLMs with conventional models, yielding two major findings.

\begin{figure*}[t]
    \centering
    \includegraphics[width=0.85\textwidth]{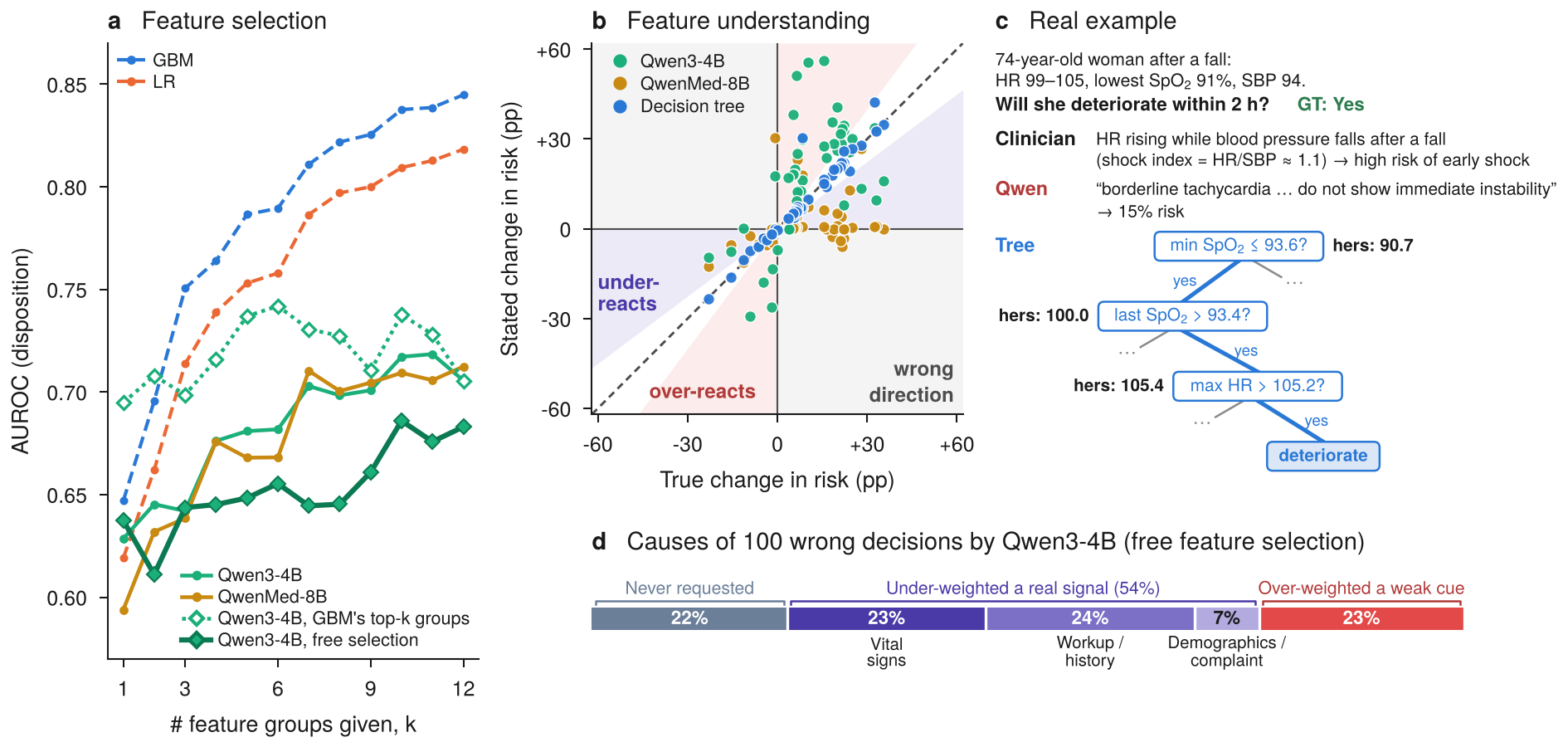}
\caption{Preliminary analysis on 2000 MC-MED patients~\citep{kansal2025mcmed}.
\textbf{(a)}~Disposition AUROC given $k$ of 12 numerical feature groups, averaged over 10 runs.
\textbf{(b)}~46 features' dataset level vs.\ model-stated change in risk. 
\textbf{(c)}~One decompensation patient.
\textbf{(d)} Error decomposition on 100 randomly sampled patients for whom the GBM predicts correctly but Qwen3-4B free selection fails, human-labeled.}
\label{fig:prelim}
\end{figure*}

First, we observe that \textbf{more EHR features do not consistently improve LLM predictions, mainly due to poor feature selection}. EHR features vary widely in format, reliability, and relevance to a particular patient outcome~\citep{shickel2018deepehr,kansal2025mcmed}.
Supervised models learn their task-specific relevance through tree splits or linear coefficients, whereas LLMs must infer relevance from context autonomously. As shown in Fig.~\ref{fig:prelim} (a), in disposition prediction, Gradient Boosting Tree Models (GBM) and Logistic Regression (LR) clearly improve as more feature groups become available, while LLMs show smaller and less consistent gains with fluctuations. Allowing the LLM to pick features like an agent selects tools to call before prediction does not resolve this problem, while supplying LLMs the top-ranked groups identified by GBM improves performance. These comparisons identify feature selection as a bottleneck for LLMs to leverage EHR features, and motivate explicit guidance on which evidence is useful for the prediction task.

Second, preliminary analysis indicates that \textbf{the LLM cannot reliably translate clinical numerical values into task-specific risk.} Even when supplied with GBM-selected features, the LLM still falls well short of GBM, raising the question why it fails to leverage these features.
For each predictive numerical feature, 
we compare the change
in model-assigned risk between patients with high and low feature values against the corresponding difference in observed outcome rates in the dataset. Fig.~\ref{fig:prelim} shows that the LLM’s risk estimates mostly fall in the over- and under-reaction region, and some in the wrong direction.
Such task-specific relationships, like the heart rate at which deterioration becomes likely, are learned from data by ML models but are not captured by LLM's medical priors~\citep{chen2026clinicalbench,moghaddam2026reprompt}. 
Error analysis in Fig.~\ref{fig:prelim}(d) further shows that among LLM prediction errors over 100 cases,  most come from mis-weighting evidence that the LLM had already retrieved. In the example shown in Fig.~\ref{fig:prelim}(c), the LLM dismisses borderline vital signs as showing no immediate instability, while tree models correctly route the same values to a high-risk prediction. This pattern reveals that even when informative evidence is provided, the LLM may misjudge its predictive significance. Effective EHR prediction therefore requires learning not only \emph{which features to select}, but also \emph{how their numerical values relate to the target outcome}.

Guided by these findings, we propose \textbf{TIR-EC}, a tool-integrated reinforcement learning (RL) framework that teaches LLMs both \emph{which} EHR features to use and \emph{how} to interpret them for EHR-based clinical task prediction. To address feature selection, we treat EHR feature groups as tools that the LLM can optionally call, so that selection becomes an action optimized with RL. Since expert tool-use trajectories are unavailable, we use tree-importance to derive a reference set of informative tools against which the agent’s selections are rewarded. To address feature understanding, we curate experience in the form of \emph{skills}, consisting of short task-specific rules each stating which evidence to inspect and how its values shift risk. We then distill these skills into the model weights through reverse curriculum learning. Training starts with the full skills in context and gradually removes them, so that the agent predicts without any skills at inference time. Experiments on disposition, decompensation, and revisit prediction show that: (1) with only a 4B backbone, TIR-EC outperforms both conventional ML models and frontier LLMs; (2) TIR-EC improves the LLM in both feature selection and feature understanding, the two weaknesses identified above; and (3) the two abilities are mutually beneficial: as the LLM understands features better, it also selects them better.

Our contributions are threefold: 
\blackcircled{1} We identify two key weaknesses behind LLMs’ underperformance on EHR prediction, poor feature selection and unreliable numerical understanding. \blackcircled{2} Based on these findings, we propose TIR-EC, reformulating feature selection as an agentic tool call task and feature understanding as a skill distillation task. \blackcircled{3} Empirical results show that TIR-EC achieves state-of-the-art performance, improves both feature selection and feature understanding, and reveals that the two are mutually beneficial.

\begin{figure*}[t]
    \centering
    \includegraphics[width=0.9\textwidth]{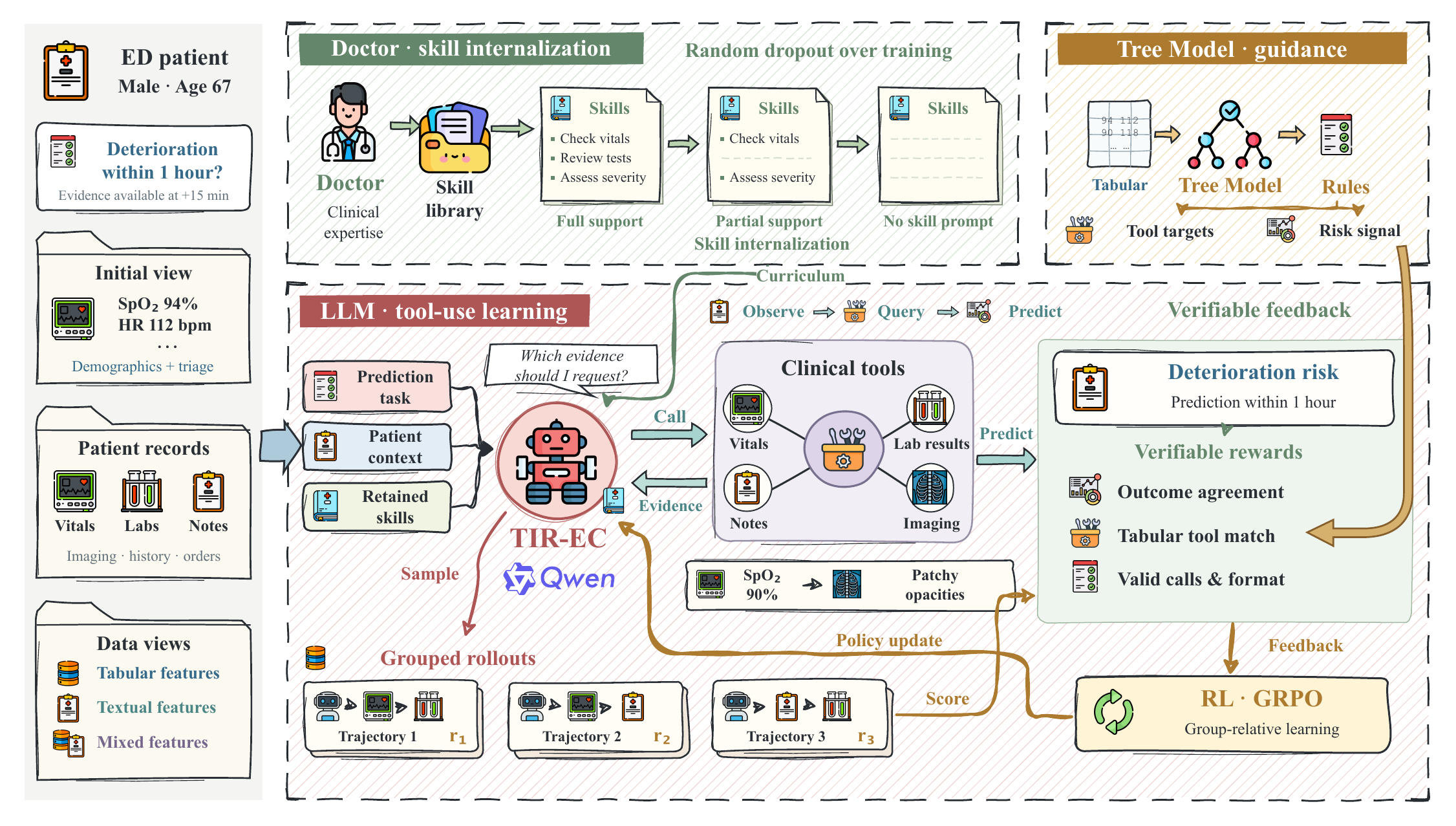}
    \caption{\textbf{Overview of TIR-EC.} The agent retrieves patient evidence through clinical tools and predicts an outcome. Tree models provide tool targets and risk signals, while clinical skill guidance fades during training. Alternative trajectories for the same patient share retained guidance and receive rewards for agentic RL updates.}
    \label{fig:framework}
\end{figure*}

\section{Related Work}

\subsection{Emergency Care Prediction}

Emergency care supports a broad range of time-sensitive clinical prediction tasks~\citep{xie2022edbenchmark, chen2023mcbec}.
Unlike longitudinal EHR tasks that aggregate medical concepts (e.g., medications) from prior visits~\citep{rasmy2021medbert, wornow2023ehrshot}, ED prediction centers on one episode and must respect when each observation becomes available.
Existing work can be categorized into two settings:
(1) \textbf{Arrival-Time Prediction:} uses patient demographics and initial vital signs to estimate critical illness, arrival triage, or resource allocations~\citep{colakca2024edtriage, sebastian2026llmtriage, solaiman2026constrainedtriage}.
(2) \textbf{Post-Arrival Prediction:} updates risk after care begins using repeated vital signs, laboratory results, orders, imaging, and physiological signals~\citep{alcaraz2025mdsed,sun2024edcopilot, kansal2025mcmed}.
Recent LLM studies in emergency care have largely focused on triage~\citep{cui2026triagereview}, where inputs are compact and consistently recorded.
Still, the later setting is more demanding for LLMs, since evidence is distributed across heterogeneous sources, includes high-frequency measurements, and requires numerical and temporal interpretation~\citep{kansal2025mcmed, alcaraz2025mdsed}.
Here, we focus on this setting by treating functions as retrieval actions over structured and unstructured clinical data.

\subsection{LLMs and Agents for EHR}

Current EHR research uses pretrained representations, retrieval, automatic feature construction, multi-agent coordination, and executable interfaces to reason over structured records~\citep{steinberg2020ehrlanguage, wornow2023ehrshot, cao2026ehrrag}.
Tool-integrated reasoning methods teach models to interleave reasoning with external actions~\citep{yao2023react, patil2023gorilla, qin2023toolllm}, while tool-integrated RL optimizes these actions from task-level feedback~\citep{li2025torl, xu2026tirjudge}.
However, most of them strengthen text reasoning around coded clinical concepts already present or retrieved, and they overlook how to teach LLMs to identify useful evidence and interpret retrieval before making predictions.
For instance, knowing heart rate matters does not tell the model what value to use or how it affects risk.
Hence, our goal is to address both by using tree models to supervise evidence selection and anomaly identification.

\section{Methodology}
\label{sec:methodology}

In this work, we introduce \method to learn which clinical evidence to retrieve and how to use it for prediction.
Figure~\ref{fig:framework} gives an overview: the agent queries clinical tools, tree-based supervision scores its evidence choices and predictions, clinical guidance fades during training, and our customized reward updates the policy from grouped rollouts.

\subsection{Task Definition}
\label{sec:task-definition}

Let $q$ identify a prediction task, $c$ denote the evidence cutoff, $h$ denote the prediction horizon, and $y\in\{0,1\}$ denote the observed outcome. The clinical data $E_{\leq c}$ is visible to the model and clinical functions.
The initial view $x_0=g_0(E_{\leq c})$ contains demographics, triage vital signs, and the chief complaint. A set of clinical functions $\mathcal{T}=\{t_1,\ldots,t_n\}$ provides selective access to the remaining record before the cutoff.
At interaction step $k$, the policy writes short reasoning $r_k$, selects a function $a_k\in\mathcal{T}$, and receives its observation $o_k$. After at most $K$ calls, it returns a probability $p\in[0,1]$ for the positive outcome. 
We denote the resulting trajectory as
\begin{equation}
    \begin{aligned}
    \tau={}&(x_0,r_1,a_1,o_1,\ldots,r_m,a_m,o_m,\\
            &\quad r_{m+1},p),\qquad m\leq K.
    \end{aligned}
    \label{eq:trajectory}
\end{equation}
In each rollout, the model writes untagged reasoning and issues a \toolcallfmt{\texttt{\textless{}tool\_call\textgreater{}}} block; the environment returns evidence in a \observationfmt{\texttt{\textless{}tool\_response\textgreater{}}} block. The final probability is enclosed in an \answerfmt{\texttt{\textless{}answer\textgreater{}}} block. Appendix~\ref{sec:tir-rollout} describes the full interaction protocol.

The policy thus controls both evidence acquisition and prediction. Training maximizes the expected terminal reward:
\begin{equation}
    \theta^\star
    = \arg\max_{\theta}\;
    \mathbb{E}_{\tau\sim\pi_\theta}
    \left[R(\tau)\right].
    \label{eq:trajectory-objective}
\end{equation}

\subsection{Reward Design and RL Training}
\label{sec:reward-design}

The outcome label indicates whether the final decision is correct, but not which evidence was useful. 
Prior to LLM training, we therefore train a GBM~\citep{natekin2013gradient} on structured features available to provide patient risk and identify task-relevant functions.
We map its feature importance to the functions that retrieve each feature family and retain those whose normalized importance exceeds $\gamma$, yielding a reference set $B_q$.
These signals are incorporated into the following reward components for policy optimization:

\paragraph{Format reward.}
$R_{\mathrm{format}}$ checks the required output and call format:
\begin{equation}
R_{\mathrm{format}}(\tau)=
\begin{cases}
1, & \substack{\text{if answer and all function}\\ 
\text{calls are parseable},}\\
0, & \text{otherwise.}
\end{cases}
\label{eq:format-reward}
\end{equation}

\paragraph{Class-safe prediction reward.}
Let $p_i^{\mathrm{tree}}$ be the calibrated risk from the gradient boosting model for case $i$. Directly copying this probability can teach the wrong class when the model is mistaken, while the hard label alone provides no graded notion of risk. We keep the tree confidence only on the label's side of the decision boundary:
\begin{equation}
\begin{aligned}
t_i &=
\begin{cases}
\max(0.5,p_i^{\mathrm{tree}}), & y_i=1,\\
\min(0.5,p_i^{\mathrm{tree}}), & y_i=0,
\end{cases}\\
R_{\mathrm{pred}}(\tau_i) &= 1-(p_i-t_i)^2.
\end{aligned}
\label{eq:class-safe-reward}
\end{equation}
Hence, an estimate on the opposite side is moved to $0.5$ instead of reversing the observed class.

\paragraph{Function matching reward.}
Let $C_i$ be the set of structured functions called in $\tau_i$. We remove functions whose information is already shown in $x_0$ and compare the remaining set with $B_q$:
\begin{equation}
R_{\mathrm{match}}(\tau_i)
=\frac{|C_i\cap B_q|}{|C_i\cup B_q|}.
\label{eq:function-match-reward}
\end{equation}
This guides which functions are useful for a task, but it does not require a sequence. Functions that return detailed values or text are excluded because the decision tree does not observe these contents.

\paragraph{Execution reward.}
$R_{\mathrm{exec}}$ checks the interaction as a whole:
\begin{equation}
R_{\mathrm{exec}}(\tau)=\mathbb{I}_{\mathrm{exec}}(\tau),
\label{eq:execution-reward}
\end{equation}
where $I_{\mathrm{exec}}(\tau)$ indicates whether the trajectory is complete and all calls are parseable, within budget, and error-free.
Unlike the format term, it depends on whether the calls execute successfully and stay within the interaction budget.

\paragraph{Total reward.}
We combine the prediction and function signals into a correctness reward $R_{\mathrm{corr}}$, then add checks for output format and execution:
\begin{equation}
\begin{aligned}
R_{\mathrm{corr}}
    &=\alpha R_{\mathrm{pred}}+(1-\alpha)R_{\mathrm{match}},\\
R(\tau)
    &=\lambda_f R_{\mathrm{format}}
      +\lambda_c R_{\mathrm{corr}}
      +\lambda_e R_{\mathrm{exec}},
\end{aligned}
\label{eq:total-reward}
\end{equation}
where $\alpha$ balances prediction and function selection, while $\lambda_f$, $\lambda_c$, and $\lambda_e$ weight the three reward groups. 
Concretely, we optimize the policy with Group Relative Policy Optimization (GRPO)~\citep{shao2024deepseekmath}.
Appendix~\ref{sec:rl-training} details the group-relative advantage and policy optimization objective.

\subsection{Skill Curriculum}
\label{sec:skill-curriculum}

For each task $q$, physicians can provide a skill document $\mathcal{S}_q$ containing general guidance for selecting and interpreting evidence. 
To internalize the skill content into the trained LLM, we use a reverse curriculum learning method with random dropout. 
At each optimizer step $s$, skill sentences are retained independently with probability
\begin{equation}
q_s=
\max\!\left(
0,\,
1-\delta\left\lfloor\frac{s}{\Delta}\right\rfloor
\right),
\label{eq:skill-schedule}
\end{equation}
where $\Delta$ is the update interval and $\delta$ is the retention, decreasing at each update. Training starts with the complete skill and removes sentences in stages until no guidance remains.

We sample one subset for the $G$ trajectories of a case, so group-relative rewards compare policy behavior under identical guidance. Validation and test prompts never contain skills; gradual removal therefore encourages the policy to internalize the guidance rather than rely on the prompt.

\section{Experiments}
\label{sec:experiments}

\subsection{Experimental Setups}

\paragraph{Dataset \& Task Formulation.}
We conduct experiments on MC-MED~\citep{kansal2025mcmed}, containing 118,385 adult emergency-department visits from 70,545 patients, using the patient-disjoint chronological split of 78\%/11\%/11\% for training, validation, and test. The dataset includes demographics, triage information, diagnoses, medications, orders, lab results, radiology reports, physiological signals, and waveform summaries. We evaluate three binary tasks: \emph{disposition}, predicting admission versus discharge at 1, 2, or 4 hours; \emph{decompensation}, predicting HR $>110$, MAP $<65$, or SpO$_2<90\%$ within 60, 90, or 120 minutes after a 15-minute assessment; and \emph{revisit}, predicting ED return within 3, 7, or 14 days after discharge. More details are shown in Appendix~\ref{app:datasets}.

\paragraph{Implementation Details.}
We use Qwen3-4B~\citep{qwen2025qwen3} without thinking mode as the backbone and train one shared policy across the three tasks. The agent selects up to 12 tool calls from 18 feature groups, with multi-turn interaction and RL implemented using verl, verl-tool, and vLLM. Skill sentences are fully retained for the first 20 steps, then their retention probability decreases by 0.05 every 20 steps until reaching zero at step 400. No skills are used for validation or test. Training uses 4 NVIDIA H100 80GB GPUs. Additional hyperparameters are provided in the Appendix~\ref{app:implement}.

\paragraph{Baselines.}
We compare TIR-EC with the following baselines: (i) \textbf{ML and DL Models}: XGBoost~\citep{chen2016xgboost}, BEHRT~\citep{li2020behrt}, MC-BEC~\citep{chen2023mcbec}, TransformerEHR~\citep{yang2023transformehr}, and TabM~\citep{gorishniy2025tabm}; (ii) \textbf{Small and Domain-Specific LLMs}: Llama 3.3 70B~\citep{meta2024llama33}, Gemma 4 31B~\citep{gemmateam2026gemma4}, GPT-OSS 120B~\citep{openai2025gptoss}, m1-7B (23K)~\citep{huang2026m1}, HuatuoGPT-o1-8B~\citep{chen2025huatuogpto1}, Baichuan-M1-14B~\citep{wang2025baichuanm1}, MedGemma-27B~\citep{sellergren2025medgemma}, Qwen3.5-27B~\citep{qwen2026qwen35}, and QwQ-32B~\citep{qwen2025qwq32b}; (iii) \textbf{Large and Frontier LLMs}: DeepSeek V4 Flash~\citep{deepseekai2026deepseekv4}, Gemini 3.7 Flash~\citep{googledeepmind2026gemini37flash}, GLM-5.2~\citep{zai2026glm52}, GPT-5.6 Sol~\citep{openai2026gpt56}, Claude Opus 5~\citep{anthropic2026opus5}, Grok 4.6~\citep{spacexai2026grok46}, Qwen 3.8 2.4T~\citep{qwen2026qwen38}, and MiniMax M2.7~\citep{minimax2026m2}; and (iv) \textbf{Backbone and Interface Controls}: Qwen3-4B~\citep{qwen2025qwen3} under both text-only and agentic settings. 

\begin{table*}[t]
  \centering
  \normalfont\color{black}
  \fontsize{7.4pt}{8.5pt}\selectfont
  \setlength{\tabcolsep}{2.0pt}
  \renewcommand{\arraystretch}{0.95}
  \resizebox{\textwidth}{!}{%
  \begin{tabular}{l|*{5}{c}|*{5}{c}|*{5}{c}|*{3}{c}}
    \toprule
    & \multicolumn{5}{c}{Disposition}
    & \multicolumn{5}{c}{Decompensation}
    & \multicolumn{5}{c}{Revisit}
    & \multicolumn{3}{c}{Macro avg.} \\
    \cmidrule(lr){2-6}
    \cmidrule(lr){7-11}
    \cmidrule(lr){12-16}
    \cmidrule(lr){17-19}
    Model
    & P & R & F1 & AUPRC & AUROC
    & P & R & F1 & AUPRC & AUROC
    & P & R & F1 & AUPRC & AUROC
    & F1 & AUPRC & AUROC \\
    \midrule
    \multicolumn{19}{l}{ML \& DL Baselines} \\
    \midrule
    XGBoost
      & 62.3 & 84.3 & 71.6 & 75.9 & 80.2
      & 27.9 & 41.7 & 33.5 & 26.9 & 65.2
      & 30.3 & 60.6 & \cellcolor{secondblue}40.4 & 26.6 & 60.9
      & 48.5 & 43.1 & 68.8 \\
    BEHRT
      & 70.2 & 76.7 & 73.3 & 82.3 & 82.8
      & 21.0 & 61.1 & 31.2 & 28.8 & 65.8
      & 24.3 & 55.1 & 33.8 & 23.7 & 57.5
      & 46.1 & 44.9 & 68.7 \\
    MC-BEC
      & 70.5 & 77.5 & 73.8 & 82.6 & 83.0
      & 20.4 & 55.6 & 29.9 & 33.0 & 64.0
      & 22.3 & 51.0 & 31.1 & 27.8 & 59.3
      & 44.9 & 47.8 & 68.8 \\
    TransformerEHR
      & 66.1 & 68.3 & 67.2 & 68.1 & 72.2
      & 24.3 & 25.0 & 24.7 & 19.6 & 53.6
      & 21.4 & 63.3 & 32.0 & 21.2 & 52.6
      & 41.3 & 36.3 & 59.5 \\
    TabM
      & 76.1 & 62.5 & 68.7 & 76.9 & 79.8
      & 19.1 & 72.2 & 30.2 & 29.7 & 66.1
      & 22.4 & 34.7 & 27.2 & 23.9 & 55.5
      & 42.0 & 43.5 & 67.1 \\
    \midrule
    \multicolumn{19}{l}{Small \& Domain-Specific LLMs} \\
    \midrule
    Llama 3.3 70B
      & 56.4 & 93.8 & 70.4 & 65.7 & 72.5
      & 40.0 & 5.6 & 9.8 & 24.1 & 58.4
      & 14.3 & 8.2 & 10.4 & 18.5 & 44.7
      & 30.2 & 36.1 & 58.5 \\
    Gemma 4 31B
      & 78.2 & 71.7 & 74.8 & 81.3 & 84.4
      & 66.7 & 11.1 & 19.1 & 29.8 & 61.7
      & 21.4 & 6.1 & 9.5 & 24.2 & 57.7
      & 34.5 & 45.1 & 67.9 \\
    GPT-OSS 120B
      & 73.9 & 78.2 & 76.0 & 83.9 & 84.4
      & 50.0 & 5.6 & 10.0 & 23.7 & 58.6
      & 28.6 & 8.3 & 12.9 & 24.3 & 59.7
      & 33.0 & 44.0 & 67.6 \\
    m1-7B (23K)
      & 48.0 & \cellcolor{bestblue}98.3 & 64.5 & 60.2 & 61.6
      & 15.5 & \cellcolor{secondblue}94.6 & 26.6 & 14.0 & 44.7
      & 22.3 & 92.7 & 36.0 & 23.9 & 47.8
      & 42.4 & 32.7 & 51.4 \\
    HuatuoGPT-o1-8B
      & 48.8 & 96.6 & 64.9 & 53.2 & 56.9
      & 20.6 & 44.6 & 28.2 & 17.2 & 54.7
      & 24.0 & 85.3 & 37.5 & 21.5 & 51.2
      & 43.5 & 30.6 & 54.3 \\
    Baichuan-M1-14B
      & 59.8 & 85.7 & 70.4 & 65.9 & 72.9
      & 26.8 & 54.1 & 35.9 & 21.0 & 61.5
      & 22.5 & 93.6 & 36.3 & 22.5 & 48.1
      & 47.5 & 36.5 & 60.8 \\
    MedGemma-27B
      & 58.9 & 84.6 & 69.4 & 65.3 & 71.1
      & 23.8 & 45.9 & 31.3 & 21.5 & 59.2
      & 22.5 & \cellcolor{bestblue}100.0 & 36.7 & 23.1 & 48.9
      & 45.8 & 36.6 & 59.7 \\
    Qwen3.5-27B
      & 62.1 & 95.1 & 75.2 & 61.5 & 72.5
      & 16.5 & 90.5 & 28.0 & 24.5 & 56.9
      & 26.2 & 67.0 & 37.6 & 27.8 & 57.1
      & 46.9 & 37.9 & 62.2 \\
    QwQ-32B
      & 58.4 & 85.9 & 69.5 & 61.9 & 68.9
      & 15.1 & \cellcolor{bestblue}100.0 & 26.2 & 19.2 & 53.6
      & 22.4 & \cellcolor{secondblue}97.2 & 36.4 & 22.5 & 50.3
      & 44.0 & 34.5 & 57.6 \\
    \midrule
    \multicolumn{19}{l}{Large \& Frontier LLMs} \\
    \midrule
    DeepSeek V4 Flash
      & 69.3 & 87.0 & 77.2 & 85.3 & 86.0
      & 62.5 & 13.9 & 22.7 & 28.8 & 59.1
      & 26.9 & 36.7 & 31.0 & 25.4 & 58.7
      & 43.6 & 46.5 & 67.9 \\
    Gemini 3.7 Flash
      & \cellcolor{secondblue}78.4 & 78.8
      & \cellcolor{secondblue}78.6 & 87.1
      & \cellcolor{secondblue}88.4
      & 40.0 & 33.3 & 36.4 & 37.4 & 64.0
      & \cellcolor{bestblue}66.7 & 4.1 & 7.7
      & \cellcolor{secondblue}33.9 & 64.8
      & 40.9 & \cellcolor{bestblue}52.8 & 72.4 \\
    GLM-5.2
      & 76.1 & 78.3 & 77.2 & 84.3 & 85.4
      & 50.0 & 11.1 & 18.2 & 28.3 & 59.2
      & 29.6 & 32.7 & 31.1 & 26.1 & 58.3
      & 42.2 & 46.2 & 67.6 \\
    GPT-5.6 Sol
      & \cellcolor{bestblue}83.5 & 71.7 & 77.1
      & \cellcolor{bestblue}88.9 & \cellcolor{bestblue}89.2
      & \cellcolor{secondblue}83.3 & 13.9 & 23.8
      & \cellcolor{secondblue}37.6 & 69.2
      & 33.3 & 2.0 & 3.9 & 31.4 & 62.5
      & 34.9 & \cellcolor{secondblue}52.6 & 73.6 \\
    Claude Opus 5
      & 71.9 & 83.3 & 77.2 & 85.7 & 86.2
      & 45.0 & 25.0 & 32.1 & 36.0 & \cellcolor{bestblue}71.2
      & 29.7 & 38.8 & 33.6
      & \cellcolor{bestblue}35.1 & \cellcolor{bestblue}67.6
      & 47.6 & 52.3 & \cellcolor{bestblue}75.0 \\
    Grok 4.6
      & 73.3 & 87.9 & \cellcolor{bestblue}79.9
      & \cellcolor{secondblue}87.8 & 88.1
      & \cellcolor{bestblue}100.0 & 8.3 & 15.4 & 33.3 & 65.6
      & 38.5 & 10.2 & 16.1 & 28.8 & 61.3
      & 37.1 & 50.0 & 71.7 \\
    Qwen 3.8 2.4T
      & 75.2 & 79.6 & 77.3 & 85.5 & 86.2
      & 50.0 & 5.6 & 10.0 & 29.2 & 62.3
      & \cellcolor{secondblue}45.5 & 10.2 & 16.7 & 28.6 & 57.4
      & 34.7 & 47.8 & 68.6 \\
    MiniMax M2.7
      & 57.6 & \cellcolor{secondblue}97.0 & 72.3 & 79.2 & 81.7
      & 50.0 & 14.3 & 22.2 & 27.5 & 59.3
      & 24.1 & 79.6 & 37.0 & 25.2 & 60.5
      & 43.8 & 44.0 & 67.2 \\
    \midrule
    \multicolumn{19}{l}{Qwen3 Backbone} \\
    \midrule
    Qwen3-4B Full Features
      & 70.5 & 60.8 & 65.3 & 69.8 & 74.0
      & 50.0 & 2.8 & 5.3 & 17.5 & 51.9
      & 21.8 & 34.7 & 26.8 & 20.8 & 53.2
      & 32.5 & 36.0 & 59.7 \\
    Qwen3-4B 
      & 66.0 & 71.3 & 68.5 & 69.0 & 73.6
      & 35.3 & 16.7 & 22.6 & 23.9 & 59.4
      & 18.3 & 22.5 & 20.2 & 21.0 & 54.6
      & 37.1 & 38.0 & 62.5 \\
    \midrule
    \multicolumn{19}{l}{\method{} Variants} \\
    \midrule

    w/o tree-guided selection
      & 74.2 & 58.8 & 65.6 & 69.9 & 75.8
      & 23.0 & 38.9 & 28.9 & 29.7 & 63.8
      & 27.7 & 46.9 & 34.9 & 29.7 & 63.4
      & 43.1 & 43.1 & 67.7 \\
    w/o teacher probability
      & 75.1 & 75.4 & 75.3 & 83.8 & 83.7
      & 33.3 & 41.7 & \cellcolor{secondblue}37.0 & 30.4 & 63.4
      & 39.0 & 32.7 & 35.6 & 29.4 & 63.4
      & \cellcolor{secondblue}49.3 & 47.9 & 70.2 \\
    w/o skill curriculum
      & 73.3 & 72.1 & 72.7 & 81.7 & 82.4
      & 26.3 & 41.7 & 32.3 & 31.7 & 67.3
      & 26.5 & 71.4 & 38.7 & 29.0 & 64.3
      & 47.9 & 47.5 & 71.3 \\
    \rowcolor{blue!4}\method{} (Qwen3-4B)
      & 69.6 & 89.6 & 78.3 & 86.7 & 87.4
      & 30.6 & 52.8
      & \cellcolor{bestblue}38.8
      & \cellcolor{bestblue}38.7
      & \cellcolor{secondblue}70.1
      & 29.9 & 71.4
      & \cellcolor{bestblue}42.2
      & 29.9 & \cellcolor{secondblue}66.4
      & \cellcolor{bestblue}53.1
      & 51.8 & \cellcolor{secondblue}74.6 \\
    \bottomrule
  \end{tabular}%
  }
  \caption{Results on three emergency-care prediction tasks. All LLMs are evaluated in the same agentic feature selection setup as TIR-EC, except Qwen's "Full Features" rows intakes all features for comparison.}
  \label{tab:main-results}
\end{table*}

\subsection{Main Results}
\textbf{At all metrics, TIR-EC beats ML and achieves best or competitive scores with large frontier LLMs with only 4B parameters}. As shown in Tab.~\ref{tab:main-results}, TIR-EC achieves the highest macro F1 of all models and AUROC second to Opus 5. Its advantage is largest on decompensation and revisit, obtaining the best F1 on both tasks and the best AUPRC on decompensation. Compared with its own backbone, TIR-EC improves F1, AUPRC, and AUROC on every task by a wide margin. Apart from TIR-EC, only four frontier LLMs (GPT-5.6 Sol, Claude Opus 5, Grok 4.6, and Gemini 3.7 Flash) surpass the ML baselines in macro AUPRC and AUROC, and all of them except Claude Opus 5 still fall behind these baselines in macro F1.

\textbf{Existing LLMs struggle with decompensation and revisit, and neither scale nor medical training fixes this.} On decompensation and revisit, the baselines fail in two opposite ways. Most general-purpose LLMs, frontier models included, are too conservative. They rarely predict a positive outcome, so their precision can be high while their recall stays very low. Many medical and reasoning-oriented LLMs do the reverse, flagging almost every patient as positive, which gives high recall but ranking quality that is often near chance. Medical specialization does not help either, as we observe that medical LLMs generally rank patients worse than general-purpose models of similar size. In contrast, TIR-EC alleviates both failure modes.

\textbf{Both the feature selection and understanding components of TIR-EC matter. }Removing either the tool-matching reward, which guides feature selection, or the skill curriculum, which targets feature understanding, lowers performance across all three tasks. Removing the GBM probability target mainly hurts ranking quality.
\section{Analysis}
\subsection{Does TIR-EC better select features?}
\label{sec:5.1}
To separate selection from interpretation, we hold the model fixed and vary only who chooses the evidence. Under a budget of $k$ feature groups, the agent requests up to $k$ groups and predicts from what it retrieved. As a control, the same model predicts from k groups drawn at random.

\begin{figure}
    \centering
    \includegraphics[width=0.8\linewidth]{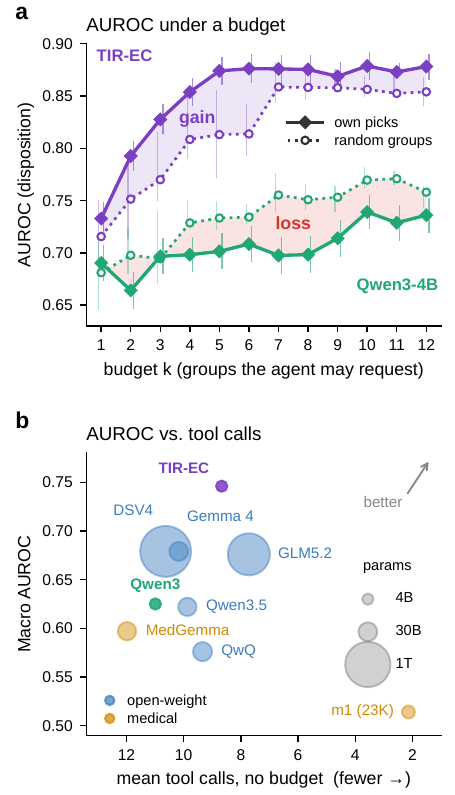}
    \caption{\textbf{Selective evidence acquisition.}
(a) Disposition AUROC when the agent may request up to $k$ feature groups (solid) or is given $k$ random groups (dotted). (b) Macro AUROC versus mean tool calls per patient.}
\label{fig:r2}
\end{figure}

\textbf{TIR-EC’s feature selection outperforms random selection at every budget, and the improvement is largest when the evidence is scarce} (low budget), as shown in Fig.~\ref{fig:r2} (a). Within only a few picks, it comes close to the accuracy it reaches with a full budget. The base model shows the reverse pattern, as its own picks fall below random selection beyond its first request.

We then remove the budget limit and compare TIR-EC with open-weight general and medical LLMs, in terms of accuracy against the number of tool calls each model makes (Fig.~\ref{fig:r2} (b)). \textbf{TIR-EC attains the highest AUROC while issuing fewer calls than most baselines; and models that call more tools are not necessarily more accurate.}

\subsection{Does TIR-EC better interpret features?}
\label{sec:5.2}

\begin{figure}
    \centering
    \includegraphics[width=0.8\linewidth]{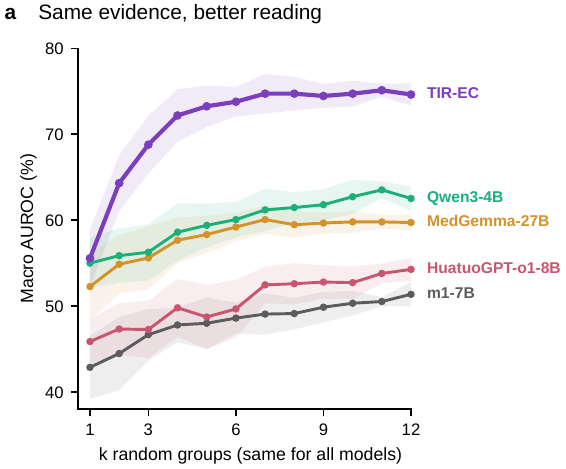}
    \caption{\textbf{Feature interpretation}. Every model receives the same $k$ randomly drawn feature groups for each patient. Lines show macro AUROC over the three tasks, averaged over 10 random draws at each $k$.}

    \label{fig:r3}
\end{figure}

To separate reading from selection, we give every model the same $k$ randomly drawn feature groups for the same patients. Any difference in accuracy then comes from how the evidence is interpreted (Fig.~\ref{fig:r3}).

With a single group, TIR-EC is no better than its base model. As groups are added, the two diverge: TIR-EC improves steeply and nears its ceiling with about half of the groups, while the base model gains only slowly. TIR-EC’s advantage lies not in reading any single measurement better, but in combining many measurements into one judgment. Medical-domain models do not show this ability.
\begin{figure}
    \centering
    \includegraphics[width=0.9\linewidth]{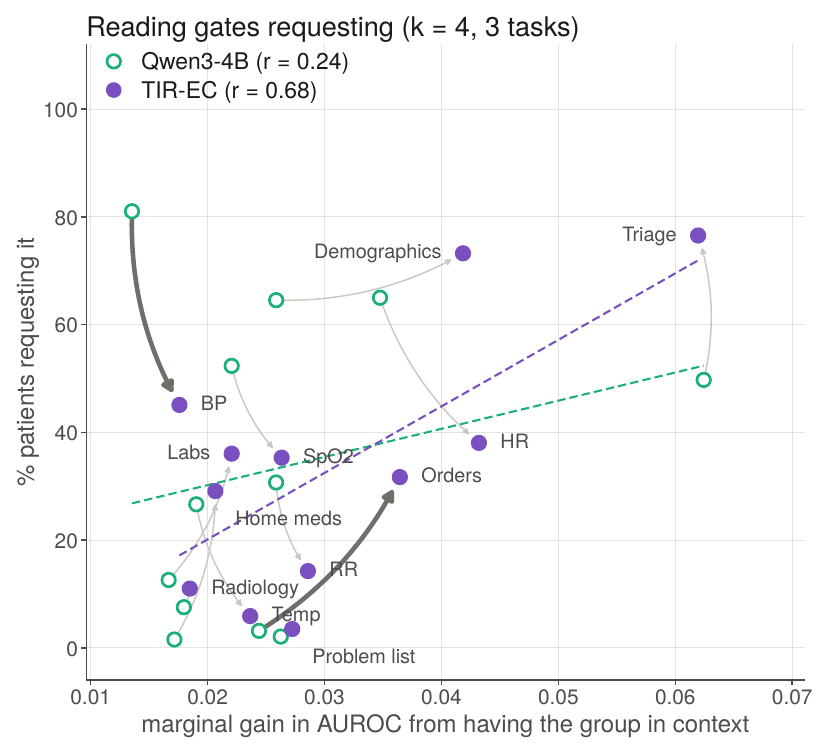}
    \caption{Feature interpretation helps selection. Each arrow is one feature group, from Qwen3-4B (tail) to \method{} (head). Only commonly selected features are shown.
Lines: least-squares fit across groups.}
\label{fig:r4}
\end{figure}
\subsection{Are feature selection and interpretation independent abilities? }
Sec.~\ref{sec:5.1} and ~\ref{sec:5.2} left two observations about the base model: its own choices are worse than random, and it extracts little from additional evidence. We ask whether these are connected. For each feature group and each model, we measure two quantities: how much the group improves that model’s own predictions when present, and how often the model requests the group under a budget of four. If selection is guided by what a model can use, the two should have a high correlation.

As shown in Fig.~\ref{fig:r4}, the base model barely requests the features it understands most. It requests blood pressure (BP) for most patients even though this is the group it benefits from least, and almost never requests orders. After training, requests closely track usable value. TIR-EC asks for the evidence it can turn into accuracy.
Training also changed what is usable. Nearly every group contributes more to TIR-EC than to its base model, and requests shifted toward the groups that gained most. As "orders" became more useful to the model, the model went from almost never requesting it to requesting it routinely. Therefore, interpretation and selection improve together: as a group becomes more useful to the agent, the agent asks for it more.

\subsection{Is TIR-EC merely mimicking the GBM teacher?}
\begin{figure}
    \centering
    \includegraphics[width=1\linewidth]{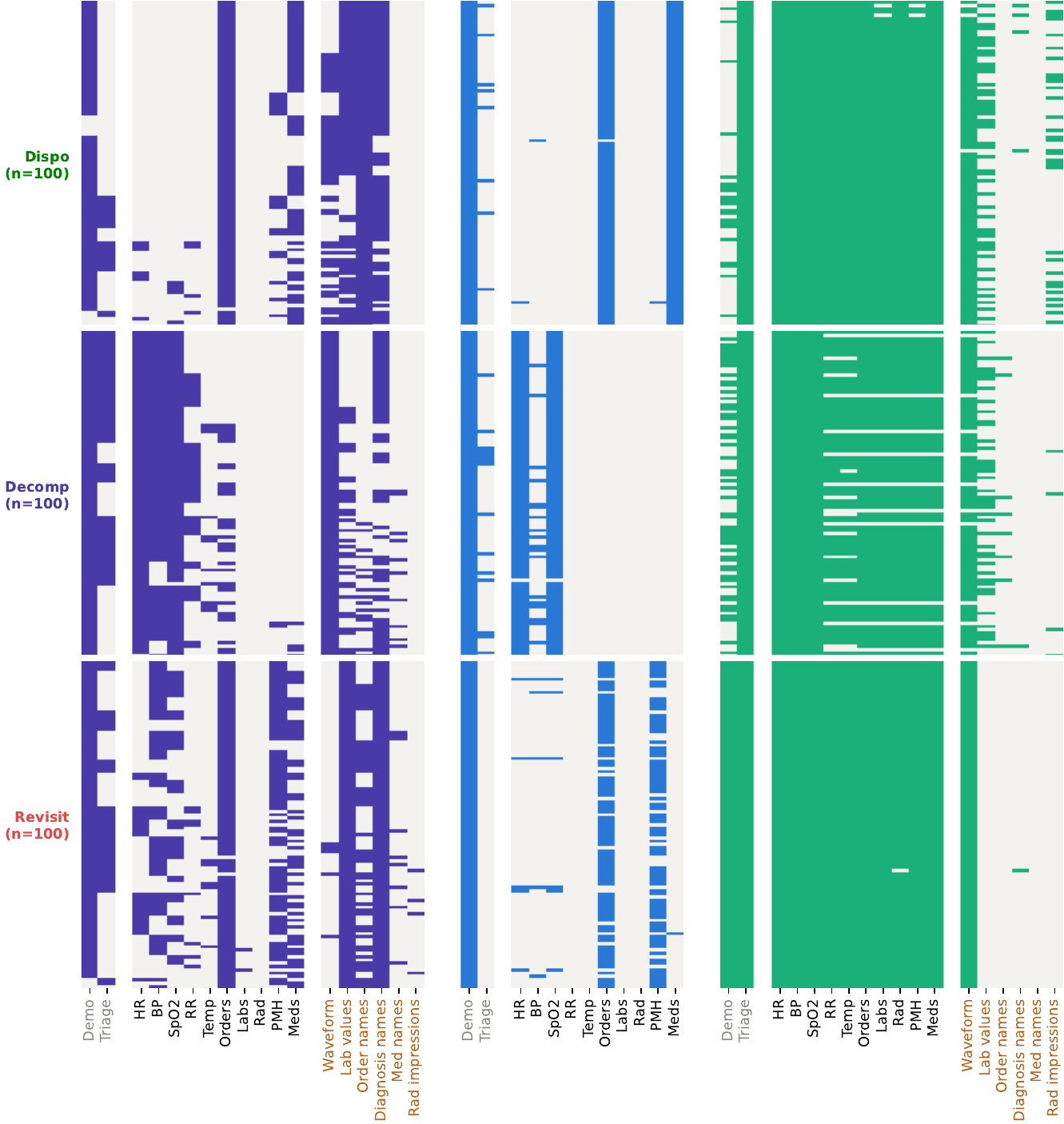}
    \caption{Barcode plot of tool call (top features for GBM) patterns for (Left) TIR-EC, (Middle) GBM, (Right) Qwen-4B base. }
    \label{fig:r5}
\end{figure}

Both of TIR-EC’s training signals come from a tree teacher. Does TIR-EC simply reproduce the teacher’s choices? To answer this, we visualize tool calls that TIR-EC, the tree, and the base model make for 100 patients per task (Fig.~\ref{fig:r5}). The five tools on the right return textual details and only the LLMs can access them.

\textbf{TIR-EC adapts its evidence to the task and to the patient: the base model asks for everything, and the tree always puts high importance on the same few groups.} TIR-EC keeps what the teacher taught it. For each task, the groups the tree relies on are among those TIR-EC requests most often. It then goes further and opens the detailed tools the tree cannot access. For disposition, for example, both rely on the orders group, but TIR-EC also reads which orders and diagnoses lie behind it. 

\subsection{Case Study}
\label{sec:paper-analysis}
How do the advantages of TIR-EC show up for individual patients? Table~\ref{tab:frontier-cases} compares TIR-EC with GPT-5.6 Sol, the strongest model on disposition, on two illustrative cases. In the first case, an older woman with an ankle fracture has one abnormal lab value and otherwise stable findings. GPT-5.6 Sol anchors on the fracture and the low hemoglobin and predicts admission. TIR-EC weighs the same abnormalities against the rest of the record, which are mostly normal labs and orders aimed at diagnosis and pain relief rather than inpatient care. It then predicts discharge, which matches the outcome. This shows that TIR-EC judges risk from the whole record rather than only from its most alarming finding, which is the feature integration ability measured in Sec.~\ref{sec:5.2}. In the second case, a man arrives after a motor vehicle crash with stable vital signs. Both models begin by checking the same vital signs and both correctly predict no deterioration. TIR-EC stops once those signs are reassuring. GPT-5.6 Sol goes on to review orders, labs, imaging, and waveforms before reaching the same answer. TIR-EC requests enough evidence rather than exhausting the feature list, illustrating the selective acquisition measured in ~\ref{sec:5.1}.

\begin{table}[t]
\centering
\begin{nolinenumbers}
\begingroup
\fontsize{8}{9.4}\selectfont
\setlength{\tabcolsep}{4pt}
\renewcommand{\arraystretch}{1.08}
\definecolor{casecream}{HTML}{F7F5ED}
\definecolor{casegpt}{HTML}{FBF1E9}
\definecolor{caseours}{HTML}{EDF5EF}
\newcommand{\caseicon}[2]{\raisebox{-2pt}{\includegraphics[height=#2]{figs/results/frontier_case_icons/#1.png}}}
\newcommand{\caseanswer}[1]{\textbf{\texttt{<answer>#1</answer>}}}
\newcommand{\casecheck}{\textcolor{teal}{\(\boldsymbol{\checkmark}\)}}
\newcommand{\casecross}{\textcolor{red!65!black}{\(\boldsymbol{\times}\)}}
\newcommand{\caseitem}[2]{\begin{minipage}[c]{\linewidth}\centering\caseicon{#1}{15pt}\par\vspace{2pt}\normalfont\fontsize{8}{9.4}\selectfont #2\end{minipage}}
\newcommand{\casegptitem}{\begin{minipage}[c]{\linewidth}\centering\includegraphics[height=15pt]{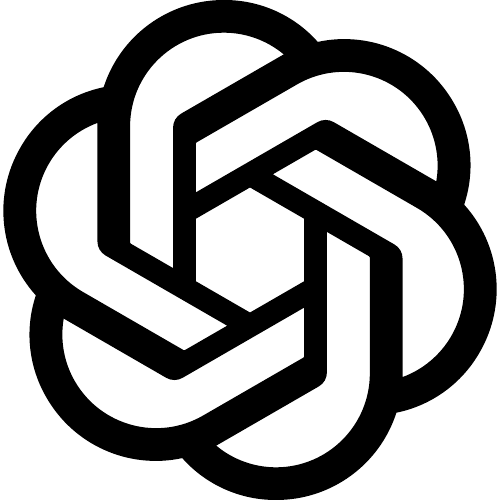}\par\vspace{2pt}\normalfont\fontsize{8}{9.4}\selectfont GPT Sol\end{minipage}}
\newcommand{\caseqwenitem}{\begin{minipage}[c]{\linewidth}\centering\includegraphics[width=34pt]{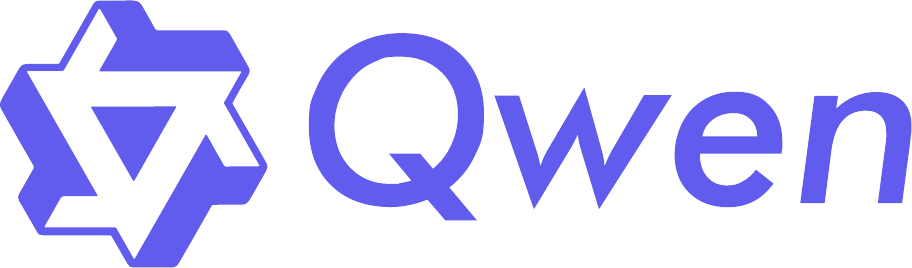}\par\vspace{2pt}\normalfont\fontsize{8}{9.4}\selectfont Our Model\end{minipage}}
\newcommand{\casecontent}[1]{\begin{minipage}[c]{\linewidth}\raggedright#1\par\vspace{3pt}\end{minipage}}
\begin{tabular}{@{}>{\centering\arraybackslash}m{44pt}>{\raggedright\arraybackslash}m{\dimexpr\columnwidth-52pt\relax}@{}}
\toprule
\rowcolor{casecream}
\multicolumn{2}{@{}p{\columnwidth}@{}}{\textbf{Case 1: Risk assessment beyond isolated abnormalities}} \\
\addlinespace[3pt]
\caseitem{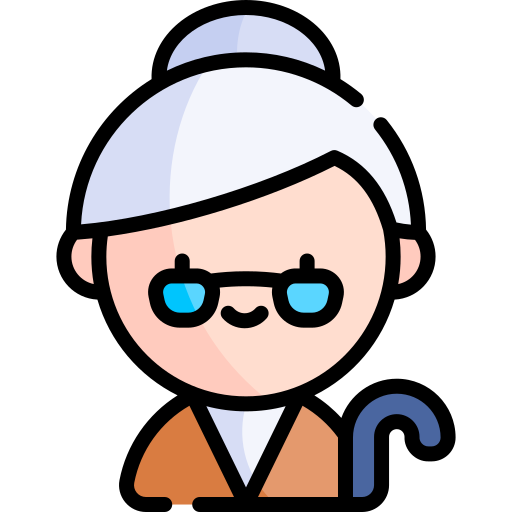}{Patient}
& \casecontent{\textbf{Woman aged 69; ankle fracture.} Predict admission at 2 h. Stable vital signs; Hb 11.4 g/dL (low); diagnostic testing and pain relief.} \\
\addlinespace[4pt]
\rowcolor{casegpt}
\casegptitem
& \casecontent{\textit{``The fracture and low hemoglobin could raise concern in an older patient. Weighting these findings above stable vital signs and supportive treatment could favor admission.''}\par\vspace{3pt}\caseanswer{0.85}\quad\casecross} \\
\addlinespace[3pt]
\rowcolor{caseours}
\caseqwenitem
& \casecontent{\textit{``While the patient has a fracture, the lab values are mostly normal [\ldots] The orders are focused on diagnostic and supportive care rather than admission-level interventions.''}\par\vspace{3pt}\caseanswer{0.29}\quad\casecheck} \\
\addlinespace[4pt]
Outcome
& \casecontent{\caseicon{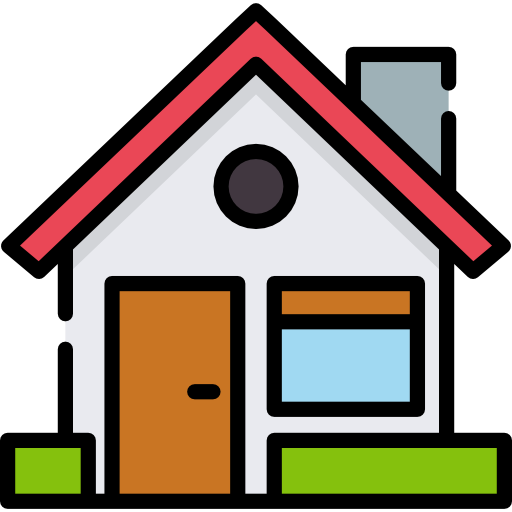}{14pt}\quad Discharged home.} \\
\addlinespace[3pt]
\midrule
\rowcolor{casecream}
\multicolumn{2}{@{}p{\columnwidth}@{}}{\textbf{Case 2: More efficient retrieval of task-relevant evidence}} \\
\addlinespace[3pt]
\caseitem{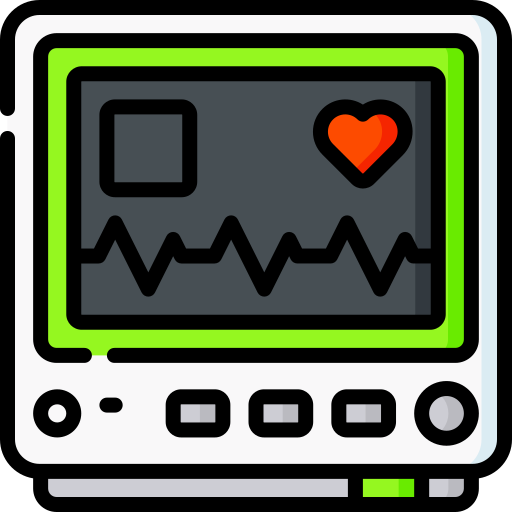}{Patient}
& \casecontent{\textbf{Man aged 46; motor vehicle crash.} Stable HR, oxygenation, and MAP; predict deterioration in 60 min.} \\
\addlinespace[4pt]
\rowcolor{casegpt}
\casegptitem
& \casecontent{\caseicon{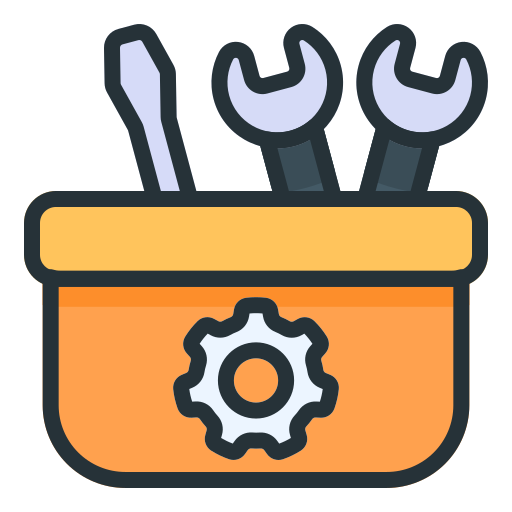}{10pt}\;\textbf{8 calls:} HR, BP, SpO\textsubscript{2}; then orders, labs, radiology, RR, and waveforms.\par\vspace{2pt}\textit{``Stable physiology could support low risk after a broader review of the trauma work-up.''}\par\vspace{3pt}\caseanswer{0.10}\quad\casecheck} \\
\addlinespace[3pt]
\rowcolor{caseours}
\caseqwenitem
& \casecontent{\caseicon{toolbox}{10pt}\;\textbf{3 calls:} HR \(\rightarrow\) SpO\textsubscript{2} \(\rightarrow\) BP \(\rightarrow\) stop.\par\vspace{2pt}\textit{``The MAP is stable, and oxygenation is adequate.''}\par\vspace{3pt}\caseanswer{0.10}\quad\casecheck} \\
\addlinespace[4pt]
Outcome
& \casecontent{\caseicon{vitals}{14pt}\quad No deterioration in the next 60 min.} \\
\addlinespace[3pt]
\bottomrule
\end{tabular}
\endgroup
\caption{Case study of TIR-EC versus GPT.}
\label{tab:frontier-cases}
\end{nolinenumbers}
\end{table}

\section{Conclusion}
We traced LLMs’ underperformance on EHR-based emergency care prediction to poor feature selection and unreliable numerical interpretation. We proposed TIR-EC, which learns to select EHR features as tools through tree-guided reinforcement learning and distills clinical skills into the model via a reverse curriculum. A 4B TIR-EC agent outperforms conventional ML baselines and is competitive with much larger frontier LLMs. Our analysis also shows that feature selection and interpretation improve together.

\section{Limitations}

\paragraph{Single-dataset evaluation.}
Our experiments are conducted on MC-MED, which provides a rich multimodal view of emergency care but represents a particular data collection setting and clinical workflow. Hospitals may differ in patient populations, measurement practices, documentation styles, and available data sources. As a result, the learned evidence-selection policy and risk interpretation may not transfer directly to other institutions. External validation across health systems is an important next step.

\paragraph{Task and label dependence.}
We study three post-arrival prediction tasks: disposition, decompensation, and revisit. These outcomes cover different stages of emergency care, but they do not represent the full range of clinical decisions. Some labels, especially disposition and revisit, can also reflect local practice patterns, resource availability, and downstream care in addition to patient physiology. Extending \method{} to other outcomes and prediction horizons would provide a broader test of its generality.

\paragraph{Dependence on tree-guided supervision.}
\method{} uses tree models to provide task-level guidance for evidence selection and graded prediction signals during training. This supervision is useful because it transfers data-driven statistical structure into the LLM, but it can also inherit dataset-specific correlations from the teacher model. In addition, our function-matching target identifies informative evidence sources at the task level rather than specifying an optimal retrieval path for every individual patient. Future work may study more adaptive or patient-specific forms of supervision.

\paragraph{Tool interface design.}
Our agent operates over a predefined set of clinical functions that expose different parts of the patient record. This design makes evidence acquisition explicit and measurable, but real EHR systems differ substantially in data schemas, tool granularity, and access constraints. Deploying the same policy in another system may therefore require remapping or redesigning the available functions, even when the underlying prediction task remains unchanged.

\paragraph{Residual numerical reasoning errors.}
Training improves how the model combines numerical evidence at the population level, but it does not guarantee correct numerical reasoning for every trajectory. Our analysis still finds occasional cases in which values are retrieved correctly but compared with a threshold in the wrong direction. Such local errors are especially important in clinical applications and suggest that learned numerical interpretation may benefit from additional verification mechanisms.

\paragraph{Retrospective evaluation.}
Our results measure predictive performance and tool-use behavior on retrospective records. They do not establish that the system improves clinician decisions, workflow efficiency, or patient outcomes when used prospectively. Clinical deployment would require prospective evaluation under real workflow constraints, including calibration, latency, alert burden, and human interaction with model outputs.

\section{Ethical Considerations}

\method{} is developed as a research framework for studying evidence selection and prediction from electronic health records. The reported results should not be interpreted as evidence that the system is ready for direct use in patient care.

\paragraph{Clinical safety.}
Incorrect predictions may lead to inappropriate escalation or false reassurance if model outputs are used without review. This risk is particularly relevant because strong aggregate performance does not eliminate individual failures, including the numerical interpretation errors observed in our trajectory analysis. Any clinical use would require independent validation, calibrated decision thresholds, appropriate human oversight, and mechanisms for detecting unreliable outputs.

\paragraph{Bias and distribution shift.}
Clinical records and outcome labels reflect both patient conditions and the health-care processes that produced them. As a result, learned prediction patterns may reproduce differences in access, treatment, documentation, or resource use across patient groups and institutions. Tree-guided supervision does not remove this issue and may transfer such correlations into the trained policy. Deployment therefore requires subgroup evaluation and continued monitoring under the target population and clinical setting.

\paragraph{Interpretability of agent behavior.}
\method{} produces observable tool calls and short natural-language reasoning, which can help inspect what evidence the model used. However, these trajectories should not be treated as faithful explanations. For clinical use, evidence provenance and verifiable checks should be preferred over relying on generated rationales alone.

\paragraph{Intended use.}
The goal of this work is to study how reinforcement learning can improve evidence acquisition and interpretation for clinical prediction. We do not recommend using \method{} as a substitute for clinician judgment or as a standalone basis for admission or discharge decisions.

\section*{Acknowledgments}

This work was supported in part by the U.S. National Science Foundation (NSF) under grants IIS-2047843, IIS-2435746, and IIS-2437621, as well as the University of Michigan MIDAS PODS Grant Award. The views and conclusions contained in this document are those of the authors and should not be interpreted as representing the official policies, either expressed or implied, of the NSF. This research was also partly supported by the Intramural Research Program of the National Institutes of Health (NIH). The contributions of the NIH author(s) are considered Works of the United States Government. This research was also partially supported by the NIH Pathway to Independence Award K99LM014903 (Q.J.). The findings and conclusions presented in this paper are those of the author(s) and do not necessarily reflect the views of the NIH or the U.S. Department of Health and Human Services.

\bibliography{custom}
\clearpage
\appendix

\begin{figure*}[t]
\centering
\begin{tcolorbox}[
    width=0.94\textwidth,
    colback=black!2,
    colframe=tirreason,
    colbacktitle=tirreason,
    coltitle=white,
    title={General Prompt Template for TIR Rollout},
    fonttitle=\bfseries\small,
    fontupper=\small,
    before upper={
        \setlength{\parindent}{0pt}
        \setlength{\parskip}{3pt}
        \setlength{\baselineskip}{11.5pt}
    },
    boxrule=0.45pt,
    arc=2mm,
    left=2mm,
    right=2mm,
    top=1.5mm,
    bottom=1.5mm
]
\textbf{Available Tools}\par
\textit{[Tool descriptions are inserted here.]} Each tool retrieves one type of clinical evidence for the current patient using only records available by the cutoff.\par

\textbf{Steps for Each Turn}\par
\textbf{1. Assess the evidence.} Review the initial patient view and all observations returned so far. Identify what evidence is still needed for the prediction.\par
\textbf{2. Decide on tool use.} If important evidence is missing, select one relevant tool. Otherwise, stop calling tools.\par
\textbf{3. Update or answer.} Interpret the new observation, revise the evidence assessment, and either continue the interaction or return the final probability.\par

\textbf{Output Format}\par
\reasonfmt{\textbf{Reasoning (untagged):}} concise evidence assessment or next-step plan.\par
\toolcallfmt{\texttt{\textless{}tool\_call\textgreater{}}}
\texttt{\{"name": "tool\_name", "arguments": \{\}\}}
\toolcallfmt{\texttt{\textless{}/tool\_call\textgreater{}}}\par
\observationfmt{\texttt{\textless{}tool\_response\textgreater{}} returned evidence
\texttt{\textless{}/tool\_response\textgreater{}}}
\textit{ (inserted by the environment)}\par
\answerfmt{\texttt{\textless{}answer\textgreater{}}$p$\texttt{\textless{}/answer\textgreater{}}}
where $0\leq p\leq1$.\par

\textbf{Important Notes}\par
Use a small number of relevant tools rather than querying every source. Treat an empty result as evidence. Emit the final answer exactly once and place no text after it.
\end{tcolorbox}
\caption{General prompt template and interaction protocol for TIR rollout. The figure summarizes the decision process rather than reproducing the runtime prompt verbatim. Blue denotes model reasoning, green a tool call, brown an environment observation, and purple the final prediction.}
\label{fig:tir-system-prompt}
\end{figure*}

\section*{Appendix}

\section{Experimental Details}

\subsection{Benchmark Datasets.}
\label{app:datasets}

We conduct our experiments on MC-MED~\citep{kansal2025mcmed}, a multi-modal emergency-care dataset containing 118,385 adult emergency-department visits from 70,545 patients. We use the patient-disjoint chronological split, with 78\%, 11\%, and 11\% of the patients assigned to the training, validation, and test sets, respectively. The available encounter data include demographics, triage information, chief complaints, prior diagnoses, home medications, orders, laboratory results, radiology reports, continuously monitored physiological measurements, and waveform-availability summaries. 
For each prediction instance, we retain only evidence observed by the task-specific cutoff. Here we consider three binary prediction tasks: \emph{disposition}, which predicts admission versus discharge at 1, 2, or 4 hours after arrival; \emph{decompensation}, which predicts whether an initially stable patient will develop heart rate above 110, mean arterial pressure below 65, or oxygen saturation below 90\% within the following 60, 90, or 120 minutes after a 15-minute assessment period; and \emph{revisit}, which predicts whether a discharged patient will return to the emergency department within 3, 7, or 14 days. Each encounter--cutoff or encounter--horizon pair is treated as one prediction instance.

\subsection{Implementation Details.}
\label{app:implement}

We use Qwen3-4B~\citep{qwen2025qwen3} without thinking mode as the backbone and train one shared policy across the three prediction tasks. 
Multi-turn tool interaction and reinforcement learning are implemented with verl, verl-tool, and vLLM. 
We sample 10,000 training instances for disposition, decompensation, and revisit.
At each optimizer step, a batch of 64 prompts generates $G=8$ trajectories per prompt, with the actor micro-batch size set to 1 per GPU. 
We train for 500 optimizer steps with a learning rate of $1\times10^{-6}$. 
The allowable prompt and response lengths are up to 6,144 and 8,192 tokens, and each trajectory may make at most 12 clinical-function calls. 
For policy optimization, we set $\epsilon_{\mathrm{low}}=0.2$, $\epsilon_{\mathrm{high}}=0.28$, the KL coefficient $\beta=10^{-4}$, and the entropy coefficient $\eta=10^{-4}$. For reward computation, we use $\alpha=0.7$ to combine prediction and function-matching rewards, with $(\lambda_f,\lambda_c,\lambda_e)=(0.1,0.8,0.1)$ for format, correctness, and execution rewards. During training, clinical-skill sentences are fully retained for the first 20 steps, after which their retention probability decreases by 0.05 every 20 steps and reaches zero after step 400. Validation and test prompts do not contain these skill sentences. We sample training trajectories with temperature 1.0 and top-$p$ 1.0, and use greedy decoding for evaluation. Checkpoints are saved every 10 steps, and the task-specific checkpoint is selected according to validation F1. All training experiments are conducted on 4 NVIDIA H100 80GB GPUs.

\section{Tool-Use Training Details}

\subsection{RL Training with GRPO}
\label{sec:rl-training}

We optimize the policy with Group Relative Policy Optimization (GRPO)~\citep{shao2024deepseekmath}. For each case, the current policy samples a group of $G$ trajectories and evaluates them with Eq.~\ref{eq:total-reward}.

\paragraph{Group-relative advantage.}
Let $\mu_R$ and $\sigma_R$ be the mean and standard deviation of the $G$ rewards for the same case. We assign trajectory $i$ the standardized advantage
\begin{equation}
A_i=
\frac{R(\tau_i)-\mu_R}{\sigma_R+\epsilon}.
\label{eq:group-advantage}
\end{equation}
This compares alternative evidence paths for one patient and avoids training a separate value model.

\paragraph{Policy objective.}
For token $z_{i,t}$ generated by the policy, let
$\rho_{i,t}=\pi_\theta(z_{i,t}\mid z_{i,<t})/
\pi_{\mathrm{old}}(z_{i,t}\mid z_{i,<t})$.
Following DAPO~\citep{yu2025dapo}, we use asymmetric clipping,
$\bar{\rho}_{i,t}=\operatorname{clip}(\rho_{i,t},1-\epsilon_l,1+\epsilon_h)$,
and aggregate the loss over generated tokens:
\begin{equation}
\begin{aligned}
\mathcal{L}_{\mathrm{RL}}
={}&-\frac{1}{\sum_{i,t}m_{i,t}}
\sum_{i,t}m_{i,t}
\min\!\left(
\rho_{i,t}A_i,\,
\bar{\rho}_{i,t}A_i
\right)\\
&+\beta\mathcal{D}_{\mathrm{KL}}-\eta\mathcal{H},
\end{aligned}
\label{eq:grpo-objective}
\end{equation}
where $m_{i,t}$ selects valid tokens generated by the model. Function observations are masked because the environment, not the policy, produces them; overlong trajectories are also masked. The upper clipping range is larger than the lower range, following DAPO, while $\beta$ and $\eta$ control KL regularization and entropy.

\subsection{TIR Rollout}
\label{sec:tir-rollout}

Figure~\ref{fig:tir-system-prompt} summarizes the interaction contract used during rollout. It is an explanatory template rather than a verbatim copy of the runtime prompt; each task inserts its outcome definition, cutoff, horizon, and tool descriptions. The interface contains summary tools for broad signals and detail tools for measured values or clinical text. The model can therefore inspect compact evidence first and request the underlying record only when needed.

The model receives $x_0$ and the function schemas, but not a serialized copy of $E_{\leq c}$. It may write a short plan in natural language and then emit one \toolcallfmt{\texttt{\textless{}tool\_call\textgreater{}}} block containing a JSON function name and an empty argument object. We disable native thinking mode, so reasoning is not wrapped in a \texttt{\textless{}think\textgreater{}} tag. Generation pauses at the closing function call tag. The environment parses the call, runs it on the record available by the cutoff, and appends the result as a user message inside \observationfmt{\texttt{\textless{}tool\_response\textgreater{}}}. The model can then revise its plan and request another source.
The interaction ends when the model emits one \answerfmt{\texttt{\textless{}answer\textgreater{}}$p$\texttt{\textless{}/answer\textgreater{}}} block or reaches the call budget $K$. Empty observations are preserved because a test or order that is absent by the cutoff can be informative. The prompt asks the model to stop early when further evidence is unlikely to change the prediction.

\onecolumn
\section{Prompts and Clinical Tools}
\label{app:prompts-tools}

This section documents the prompts and tool interface used by \method. We separate the shared instructions, task definitions, training-time skills, and tool catalog. The prompt cards consolidate repeated instructions for readability while preserving the task definitions, numerical thresholds, and output contract. Bracketed fields are filled for each case; they are not patient examples. Appendix~\ref{sec:tir-rollout} describes how these components form a multi-turn interaction.

\begingroup
\raggedbottom
\newtcolorbox{tirconfig}[2][tirreason]{
    enhanced,breakable,
    colback=#1!3,colframe=#1,colbacktitle=#1,coltitle=white,
    title={#2},fonttitle=\bfseries,fontupper=\small,
    boxrule=0.45pt,arc=1mm,
    left=2.5mm,right=2.5mm,top=2mm,bottom=2mm,
    before skip=7pt,after skip=7pt,
    before upper={\setlength{\parindent}{0pt}\setlength{\parskip}{3pt}\setlength{\baselineskip}{11.5pt}},
    title after break={#2 (continued)}
}
\newcommand{\tirslot}[1]{\textit{[#1]}}
\newcommand{\tirfn}[1]{\texttt{\detokenize{#1}}}

\subsection{Shared Prediction Prompt}
\label{app:shared-prompt}

The system message combines shared evidence-use instructions with one task definition from Appendix~\ref{app:task-prompts}. The chat template supplies the function names, descriptions, and argument schemas. The user message supplies the cutoff, horizon, and initial patient view, followed by the probability-output instruction. During training, the retained skill sentences are appended to this user message; evaluation omits them.

\begin{tirconfig}{Shared System Instructions}
You evaluate an emergency-department encounter using the evidence available at the specified cutoff. You receive an initial patient view, a prediction task, and clinical tools for requesting additional evidence.

\textbf{Available Tools}\par
\tirslot{Function names, descriptions, and schemas from Appendix~\ref{app:tool-catalog}.}
Each tool has a single purpose and takes no arguments. The environment supplies the current case and cutoff.

\textbf{Evidence Use}\par
Reason only from the initial view and observations returned by the tools. Do not invent measurements, findings, or events. Treat an empty response as informative rather than filling in missing values. Use the evidence available by the cutoff to predict the specified outcome; do not treat a future event as an observed fact.

\textbf{Steps for Each Turn}\par
1. Review the available evidence and identify what is still needed.\par
2. Select a relevant tool, or stop if further evidence is unlikely to change the prediction. Prefer a few targeted calls to an exhaustive search.\par
3. Interpret the returned evidence before deciding whether to call another tool or answer. Do not repeat the same unsuccessful or empty call.

\textbf{Interaction and Output}\par
You may write short, untagged reasoning between calls. Make at most \tirslot{tool budget} calls; the experiments use a budget of 12. Keep a turn available for the final answer.\par
\toolcallfmt{\texttt{<tool\_call>}}\texttt{\{"name": "tool\_name", "arguments": \{\}\}}\toolcallfmt{\texttt{</tool\_call>}}\par
The environment returns the observation inside \observationfmt{\texttt{<tool\_response>... </tool\_response>}}.\par
Finish with exactly one \answerfmt{\texttt{<answer>}$p$\texttt{</answer>}}, where $p\in[0,1]$ is the probability of the positive outcome. Put no text after this block.
\end{tirconfig}

\begin{tirconfig}{Initial User Message}
\textbf{Case Context}\par
Case ID: \tirslot{identifier}\quad Cutoff: \tirslot{timestamp}\quad Horizon: \tirslot{task horizon}

\textbf{Initial Patient View}\par
Demographics: \tirslot{age, gender, race, payor, and arrival mode}\par
Triage Vitals: \tirslot{temperature, HR, RR, BP/MAP, SpO$_2$, and acuity}\par
Chief Complaint: \tirslot{presenting complaint}

\textbf{Prediction Request}\par
No external prediction has been provided. Use the available tools to gather additional evidence, then return the probability of \tirslot{admission / decompensation / revisit}. Write a decimal number between 0 and 1 inside the final \answerfmt{\texttt{<answer>}} block, not a category word.
\end{tirconfig}

\clearpage
\subsection{Task-Specific Prompts}
\label{app:task-prompts}

The following task definitions accompany the shared instructions. The probability requirement takes precedence over categorical answer examples in the underlying task templates. Native thinking mode is disabled: the model may still reason in ordinary text, but no \texttt{<think>} block is required.

\begin{tirconfig}{Disposition Prediction}
\textbf{Task}\par
Predict whether this patient will be admitted from the emergency department or discharged home, using evidence available at the cutoff.

\textbf{Cutoff and Outcome}\par
The cutoff is 1, 2, or 4 hours after arrival, as specified in the user message. Admission includes an inpatient bed, observation, or ICU. Discharge means being sent home directly from the emergency department.

\textbf{Output}\par
Return the probability of admission as \answerfmt{\texttt{<answer>}$p$\texttt{</answer>}}.
\end{tirconfig}

\begin{tirconfig}{Decompensation Prediction}
\textbf{Task}\par
Predict an acute decompensation event in the specified window after the cutoff.

\textbf{Cutoff and Outcome}\par
The cutoff is 15 minutes after room assignment. The prediction window is the following 60, 90, or 120 minutes. An event is the first occurrence of any of: heart rate $>110$ beats/min, mean arterial pressure $<65$ mmHg, or SpO$_2<90\%$.

\textbf{Interpretation}\par
These thresholds define the future outcome, not a test of whether the patient is unstable at the cutoff. Forecast from the observed physiological trajectory, treatments, and clinical state.

\textbf{Output}\par
Return the probability of decompensation within the stated window as \answerfmt{\texttt{<answer>}$p$\texttt{</answer>}}.
\end{tirconfig}

\begin{tirconfig}{Revisit Prediction}
\textbf{Task}\par
Predict whether a discharged patient will return to the emergency department within the specified follow-up window.

\textbf{Cutoff and Outcome}\par
The cutoff is departure from the index visit. The follow-up window is 3, 7, or 14 days after departure. The positive outcome is a new emergency-department visit within that window.

\textbf{Interpretation}\par
Use the index encounter and available history. New events after discharge are not available. The absence of acute findings at discharge does not by itself rule out a revisit; unresolved symptoms and other incompletely observed factors may still matter.

\textbf{Output}\par
Return the probability of revisit as \answerfmt{\texttt{<answer>}$p$\texttt{</answer>}}.
\end{tirconfig}

\paragraph{Text-only comparison.}
The text-only baseline replaces tool access with clinical data placed directly in the user message. Its instructions state that no tools are available and ask the model to reason over the supplied record before returning the task prediction. This changes how evidence is provided; it is not an additional component of \method.

\clearpage
\subsection{Training-Time Skill Prompts}
\label{app:skill-prompts}

The curriculum supplies disposition and decompensation skills, but no additional revisit skill. These cards group the runtime sentences by purpose. During training, sentences are sampled independently, kept in order, and shared across a case's rollout group. Evaluation uses no skills. The numerical rules are benchmark-specific training guidance, not clinical recommendations or definitions of the outcome.

\begin{tirconfig}[teal]{Disposition Skill}
\textbf{Select Evidence}\par
Request \tirfn{get_order_counts} for the total and category breakdown of orders, and \tirfn{get_med_counts} for home-medication burden. Use the triage values already shown instead of requesting them again.

\textbf{Interpret the Pace of Care}\par
Interpret order counts relative to the time since arrival. A typical total is roughly 12 orders at one hour and 20 at four hours. Calculate orders per hour: a moderate total at four hours represents a slow work-up and leans toward discharge, whereas the same total at one hour is less informative.

\textbf{Combine Counts and History}\par
More than approximately five medication orders together with more than 34 total orders leans toward admission. More than approximately 30 total orders also leans toward admission without home medications.\par
The presence of home medications and more than approximately four orders per hour leans toward admission even when the total is small. No home medications, at most approximately 22 total orders, and at most four laboratory orders leans toward discharge.\par
A greater home-medication burden leans toward admission; older age provides only a mild upward adjustment.

\textbf{Read Order Categories}\par
Explicitly identify Code Status, Blood Bank, Nursing Transfusion, or REHAB categories when present; the skill treats them as strong admission signals. A Diet order is a weak signal and should receive little weight.

\textbf{Keep Claims Grounded}\par
Do not base disposition primarily on triage vital signs. Never infer disposition from race or ethnicity. Support severity claims with retrieved counts or categories; state plainly when the work-up is unremarkable for the horizon.
\end{tirconfig}

\begin{tirconfig}[teal]{Decompensation Skill}
\textbf{Select Evidence}\par
Request \tirfn{get_vital_hr}, \tirfn{get_vital_spo2}, \tirfn{get_vital_bp}, and \tirfn{get_vital_rr} when the corresponding measurements are needed. These tools return minimum, maximum, mean, and last values. Inspect peak HR, lowest SpO$_2$, lowest MAP, and peak RR: averages can hide abnormal readings.

\textbf{Compare Values Explicitly}\par
Before calling a value abnormal, state the measurement and its reference range. The skill uses HR outside 60--100 beats/min, SpO$_2<94\%$, RR outside 12--20 breaths/min, MAP $<65$ mmHg, SBP $<100$ mmHg, or temperature outside 36--38~$^\circ$C. Do not use terms such as hypoxia, tachycardia, or hypotension without a supporting comparison.

\textbf{Combine Physiological Signals}\par
Low risk: peak HR $\leq95$, lowest SpO$_2\geq94\%$, and lowest MAP $>77$ mmHg. The skill identifies this group as common and below the overall event rate.\par
Elevated risk: peak HR $>100$ or lowest SpO$_2<94\%$, together with at least two abnormalities among HR $>100$, SpO$_2<94\%$, MAP $<70$ mmHg, and RR $>22$.\par
Highest risk: the elevated-risk conditions hold, and mean HR $>102$ or lowest SpO$_2\leq92\%$.

\textbf{Avoid Unsupported Shortcuts}\par
ESI 2 alone is not a high-risk finding. Do not infer physiological risk from monitoring volume, race, or ethnicity. Normal current vital signs imply lower risk, not zero future risk; the outcome concerns the window after the cutoff.

\textbf{Keep Claims Grounded}\par
Support each risk claim with a retrieved value. State plainly when all measurements fall within their reference ranges.
\end{tirconfig}

\clearpage
\subsection{Clinical Tool Catalog and Responses}
\label{app:tool-catalog}
\small

The interface exposes 18 functions: 12 structured-summary functions and 6 functions for detailed values or text. All use an empty argument object. The environment binds the case identifier, cutoff, and task, so the model chooses an evidence source rather than writing a database query. Detailed-value and text functions are excluded from the function-matching reward. Demographics and triage measurements are already present in the initial view and are also excluded from that reward comparison.

\begin{tirconfig}[tirtool]{Structured-Summary Tools}
\begin{tabular}{@{}p{0.29\linewidth}p{0.67\linewidth}@{}}
\textbf{Function} & \textbf{Returned Evidence} \\
\hline
\tirfn{get_vital_hr} & Heart-rate minimum, maximum, mean, last value, and number of measurements. \\[3pt]
\tirfn{get_vital_spo2} & The same statistics for oxygen saturation. \\[3pt]
\tirfn{get_vital_bp} & Separate statistics for mean, systolic, and diastolic blood pressure. \\[3pt]
\tirfn{get_vital_rr} & Respiratory-rate statistics. \\[3pt]
\tirfn{get_vital_temp} & Temperature statistics. \\[3pt]
\tirfn{get_triage_vitals} & Triage temperature, HR, RR, SpO$_2$, SBP, DBP, and acuity. \\[3pt]
\tirfn{get_order_counts} & Total number of orders and counts by order type. \\[3pt]
\tirfn{get_lab_counts} & Number of laboratory component results, distinct panels, and results with an abnormality flag. \\[3pt]
\tirfn{get_pmh_counts} & Total prior-history records, number of top categories shown, and number of recent distinct categories in the history digest. \\[3pt]
\tirfn{get_med_counts} & Total active home-medication records, number of classes shown, and number of recent distinct medications in the digest. \\[3pt]

\tirfn{get_rad_counts} & Number of radiology studies and number with an impression. \\[3pt]
\tirfn{get_demographics} & Age, gender, race, ethnicity, arrival mode, payor, and chief complaint. \\
\end{tabular}
\end{tirconfig}

\begin{tirconfig}[tirtool]{Detailed-Value and Text Tools}
\begin{tabular}{@{}p{0.29\linewidth}p{0.67\linewidth}@{}}
\textbf{Function} & \textbf{Returned Evidence} \\
\hline
\tirfn{get_waveform_intensity} & Precomputed waveform duration and segment count for each channel type; no raw waveform signal. \\[3pt]
\tirfn{get_lab_values} & Up to 24 most recent results: panel, component, value, units, abnormality flag, and result time. \\[4pt]
\tirfn{get_orders_detail} & Up to 15 most frequent procedure names and their counts. \\[4pt]
\tirfn{get_diagnoses_detail} & Up to 10 frequent prior comorbidity categories with counts, followed by up to 10 additional recent distinct categories. \\[4pt]
\tirfn{get_medications_detail} & Up to 10 recent distinct home medications and their medication classes. \\[4pt]
\tirfn{get_rad_impressions} & Radiology study names, de-identified impression text, and result times. \\
\end{tabular}
\end{tirconfig}

\paragraph{Time and content restrictions.}
Timestamped records are filtered by availability: physiological measurements use measurement time, orders use order time, and laboratories and radiology use result time. Prior diagnoses and home medications use their recorded dates; home medications must be active at the cutoff. Final disposition, lengths of stay, and subsequent-visit outcomes are not returned. For disposition, explicit admission, bed-request, and ward-transfer orders are filtered from the order tools. Waveform availability is a precomputed encounter-level summary without per-segment time anchors, rather than a time-resolved signal.

\paragraph{Observation rendering.}
Results are rendered as key--value pairs, a summary table, or a table of records. A response header identifies the function, underlying row count, and cutoff. Detail extractors can retain fewer rows than this header count; the generic record renderer displays at most 50 rows. Empty results remain visible. The example below shows the interface with placeholders, not a real patient or a measured prediction.

\begin{tirconfig}[tirtool]{Tool Schema and Observation Format}
\begin{minipage}[t]{0.47\linewidth}
\textbf{Schema Supplied to the Model}\par
Name: \tirfn{get_vital_hr}\par
Description: Heart-rate aggregates (min/max/mean/last/n) over (arrival, cutoff].\par
Parameters: \texttt{\{"type": "object", "properties": \{\}\}}
\end{minipage}\hfill
\begin{minipage}[t]{0.49\linewidth}
\textbf{Environment Observation (Placeholders)}\par
\observationfmt{\texttt{<tool\_response>}}\newline
\texttt{get\_vital\_hr (rows: [n],}\newline
\texttt{cutoff: [timestamp])}\newline
\texttt{measure\ min\ max\ mean\ last\ n}\newline
\texttt{HR\quad [min]\ [max]\ [mean]\ [last]\ [n]}\newline
\observationfmt{\texttt{</tool\_response>}}
\end{minipage}
\end{tirconfig}
\endgroup

\end{document}